\documentclass[letterpaper, 10 pt, conference]{ieeeconf}  %

\IEEEoverridecommandlockouts                              %

\usepackage{graphics} %
\usepackage{epsfig} %
\usepackage{mathptmx} %
\usepackage{times} %
\usepackage{amsmath} %
\usepackage{amssymb}  %

\usepackage{cuted}   %
\usepackage{capt-of} %

\usepackage{cite}
\usepackage{url}
\usepackage[pagebackref=false,breaklinks=true,colorlinks,bookmarks=true,bookmarksnumbered=true]{hyperref}
\usepackage{prettyref} %

\newrefformat{fig}{Fig.~\ref{#1}}
\newrefformat{sec}{Sec.~\ref{#1}}
\newrefformat{tab}{Tab.~\ref{#1}}
\newrefformat{algo}{Algo.~\ref{#1}}
\newrefformat{eq}{Eq.~\ref{#1}}

\usepackage[percent]{overpic}

\usepackage{xcolor}
\usepackage{adjustbox}

\usepackage{tikz}
\usetikzlibrary{shapes.geometric, arrows.meta, positioning, calc, fit, backgrounds}

\usepackage{booktabs}
\usepackage{multirow}

\title{\LARGE \bf
Towards Human-Like Manipulation through RL-Augmented Teleoperation and Mixture-of-Dexterous-Experts VLA
}

\author{Tutian Tang$^{13*}$, Xingyu Ji$^{21*}$, Wanli XING$^{2*}$, Ce Hao$^{4}$,\\
Wenqiang Xu$^{5}$, Lin Shao$^{5}$, Cewu Lu$^{16}$, Qiaojun Yu$^{3\dagger}$, Jiangmiao Pang$^{3\dagger}$ and Kaifeng Zhang$^{2\dagger}$%
\thanks{$^{*}$ These authors contribute equally to this work. $^{\dagger}$ Corresponding authors.}%
\thanks{$^{1}$ Shanghai Jiao Tong University $^{2}$ Sharpa $^{3}$ Shanghai AI Lab}
\thanks{$^{4}$ Zhongguancun Academy $^{5}$ National University of Singapore}
\thanks{$^{6}$ Shanghai Innovation Institute}
}

\begin{document}
\bstctlcite{IEEEexample:BSTcontrol}

\maketitle
\thispagestyle{empty}
\pagestyle{empty}

\begin{strip}
\centering
\vspace{-2cm} %
    \includegraphics[width=\linewidth]{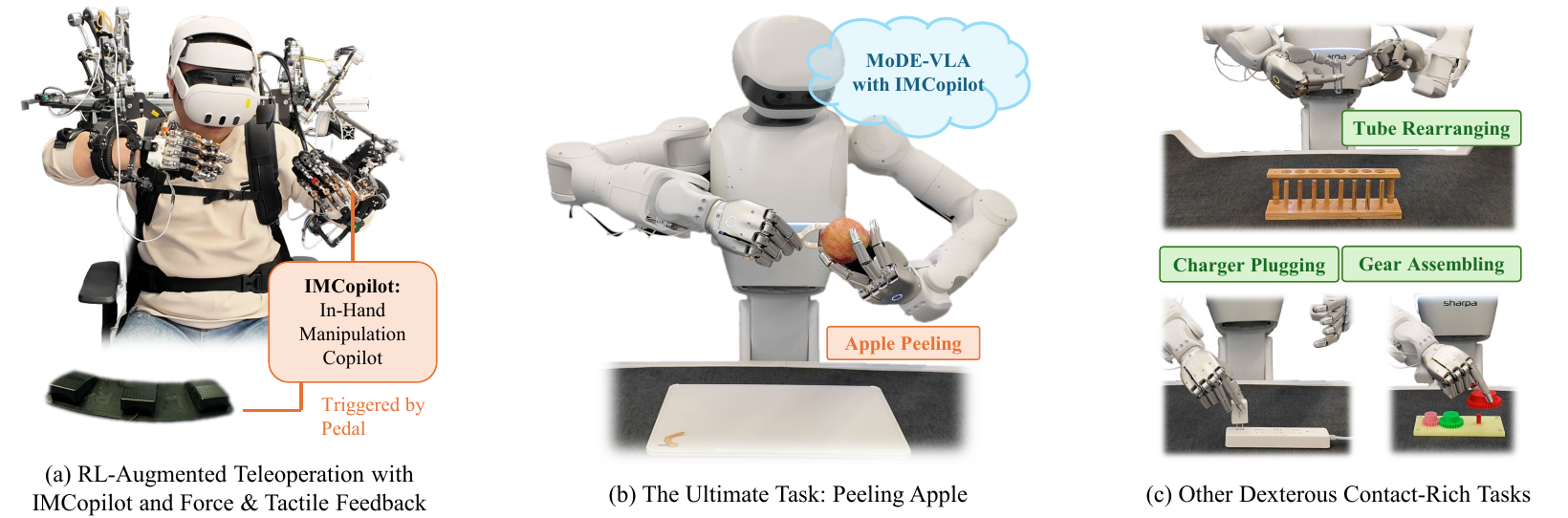}
    \captionof{figure}{
    Overview of our proposed framework. (a) We introduce an RL-augmented teleoperation system equipped with force and tactile feedback, featuring the IMCopilot to assist human operators. (b) With data collected, we train the MoDE-VLA model capable of executing highly complex, long-horizon tasks such as peeling an apple. Here IMCopilot works with VLA as a callable low-level skill for in-hand manipulation. (c) Our learned policy successfully generalizes to a variety of other dexterous, contact-rich tasks, including tube rearranging, charger plugging, and gear assembling.
}
    \label{fig:teaser}
\end{strip}

\begin{abstract}

While Vision-Language-Action (VLA) models have demonstrated remarkable success in robotic manipulation, their application has largely been confined to low-degree-of-freedom end-effectors performing simple, vision-guided pick-and-place tasks. Extending these models to human-like, bimanual dexterous manipulation—specifically contact-rich in-hand operations—introduces critical challenges in high-fidelity data acquisition, multi-skill learning, and multimodal sensory fusion. In this paper, we propose an integrated framework to address these bottlenecks, built upon two components. First, we introduce IMCopilot (In-hand Manipulation Copilot), a suite of reinforcement learning-trained atomic skills that plays a dual role: it acts as a shared-autonomy assistant to simplify teleoperation data collection, and it serves as a callable low-level execution primitive for the VLA. Second, we present MoDE-VLA (Mixture-of-Dexterous-Experts VLA), an architecture that seamlessly integrates heterogeneous force and tactile modalities into a pretrained VLA backbone. By utilizing a residual injection mechanism, MoDE-VLA enables contact-aware refinement without degrading the model's pretrained knowledge. We validate our approach on four tasks of escalating complexity, demonstrating doubled success rate improvement over the baseline in dexterous contact-rich tasks.

\end{abstract}

\section{Introduction}
\label{sec:intro}

Dexterous manipulation in humans arises from the seamless integration of hierarchical decision-making, rich tactile feedback, and well-practiced manipulation skills~\cite{johansson2009coding, merel2019hierarchical}. 
For example, when peeling an apple, humans coordinate both hands while using vision to guide the peeler, fingertip tactile sensing to detect slippage and regulate grip force, and in-hand manipulation skills to rotate the apple between successive cuts. 
Such capabilities highlight the importance of combining multimodal sensing with hierarchical manipulation skills for executing contact-rich interactions. 
Replicating these abilities in robots requires not only high-degree-of-freedom (DoF) hardware, but also learning frameworks that acquire hierarchical manipulation skills grounded in multimodal sensory feedback.

To enable robots to acquire such capabilities, recent work has explored Vision-Language-Action (VLA) models~\cite{black2024pi_0, kim2024openvla, zitkovich2023rt}, which leverage large-scale vision and language pretraining to learn general-purpose manipulation policies. However, the success of existing VLAs has largely been confined to low-DoF end-effectors such as parallel-jaw grippers~\cite{black2024pi_0, pi0_5}, where manipulation is typically reduced to simple pick-and-place primitives that rely primarily on vision rather than force or tactile sensing. This paradigm breaks down when manipulation transitions from \emph{grasping} to \emph{in-hand manipulation}, where fingers must continuously regulate contact states to reposition, rotate, or manipulate objects and tools.

Extending VLA models to human-like bimanual dexterous manipulation introduces three key challenges.
First, a severe \emph{data acquisition bottleneck}: teleoperating a bimanual system with 63 DoFs is highly challenging, and collecting high-quality demonstrations for precise multi-finger coordination in in-hand manipulation often exceeds the capabilities of even expert operators~\cite{qin2023anyteleop, cheng2024open, qin2022dexmv, an2025dexterous}.
Second, the \emph{multi-skill learning challenge}: tasks such as apple peeling span vision-guided gross motions (approaching and aligning), force-guided fine motions (cutting), and tactile-guided in-hand routines (rotating the apple). These distinct phases require different skills, making them difficult for a single policy to master within a high-dimensional action space~\cite{chi2025diffusion, zhao2023learning}.
Third, the \emph{modality heterogeneity challenge}: directly concatenating force and tactile observations into a pretrained VLA backbone fails to account for their different temporal dynamics and physical semantics, often degrading rather than improving performance~\cite{yu2025forcevla, higuera2024sparsh, liu2025forcemimic}.

To address these challenges, we propose an integrated framework for human-like bimanual dexterous manipulation built on two synergistic components.
First, \textbf{IMCopilot} (\textbf{I}n-hand \textbf{M}anipulation \textbf{Copilot}) is a set of RL-trained atomic in-hand manipulation skills that serves a \emph{dual role}.
During data collection, it functions as a shared-autonomy copilot~\cite{reddy2018shared}: the operator performs gross motions via exoskeleton teleoperation while delegating challenging in-hand manipulation phases to IMCopilot through foot pedals, enabling efficient acquisition of high-fidelity demonstrations.
During autonomous execution, IMCopilot acts as a callable low-level skill invoked by the VLA for in-hand manipulation, forming a hierarchical architecture analogous to human motor control.
Second, \textbf{MoDE-VLA}, a \textbf{M}ixture-\textbf{o}f-\textbf{D}exterous-\textbf{E}xperts VLA, addresses modality heterogeneity by introducing a dedicated pathway for force and tactile inputs.
Force–tactile tokens interact with backbone representations through self-attention, are refined via sparse MoE routing~\cite{shazeer2017outrageously} for per-timestep expert specialization, and are injected as residual corrections, enabling contact-aware refinement while preserving pretrained knowledge.

Our contributions are summarized as follows:
\begin{enumerate}
    \item \textbf{IMCopilot}: an RL-trained in-hand manipulation primitive suite that unifies teleoperation assistance and autonomous low-level control, yielding a hierarchical framework in which the VLA handles vision-language-guided planning while IMCopilot provides reactive in-hand dexterity.

    \item \textbf{MoDE-VLA}: a Mixture-of-Dexterous-Experts VLA that fuses force and tactile modalities into a pretrained VLA backbone through dedicated self-attention, sparse expert routing, and residual injection, enabling contact-aware action generation without degrading the backbone's pretrained capabilities.

    \item \textbf{Experimental validation}: we evaluate MoDE-VLA on four tasks of escalating contact complexity (gear assembling, charger plugging, test tube rearranging, and apple peeling) and demonstrate significant improvements over baselines. To our knowledge, we present the first autonomous dual-dexterous-hand apple peeling, a task requiring the full synergy of all proposed components.
\end{enumerate}

\section{Related Work}
\subsection{Teleoperation System for Dexterous Manipulation}
Teleoperation is fundamental to data-driven dexterous manipulation, providing demonstrations for learning high-DoF, contact-rich in-hand manipulation skills, and existing data-collection pipelines can be broadly categorized into \emph{vision-based}, \emph{glove-based}, and \emph{exoskeleton-based} approaches.
\textbf{Vision-based teleoperation} leverages cameras to estimate human hand pose and retarget it to robotic hands~\cite{li2019vision, qin2023anyteleop, jin2024vision}. While lightweight and scalable, such systems suffer from occlusion and depth ambiguity, and rely primarily on kinematic retargeting without explicit force or contact awareness, limiting their effectiveness in contact-rich manipulation.
\textbf{Glove-based teleoperation} employs data gloves with bend sensors or inertial units to directly measure finger joint configurations, improving retargeting accuracy and temporal consistency over vision-based methods~\cite{qin2022dexmv, cheng2024open, wang2024dexcap, gao2025glovity, mizera2019evaluation}. 
However, despite enabling scalable and high-fidelity kinematic capture, the lack of force feedback and human–robot morphological mismatch still limit their applications in contact-rich manipulation.
\textbf{Exoskeleton-based teleoperation} interfaces mechanically couple human and robot joints, reducing retargeting error and enabling bidirectional force feedback~\cite{an2025dexterous, zhang2025doglove, du2025mile}. While this tight embodiment favors contact-rich manipulation, such systems are complex and cognitively demanding, and precise multi-finger in-hand coordination often exceeds human teleoperation capability, leading to inconsistent demonstrations. In summary, existing pipelines leave the most challenging contact-rich in-hand manipulation to human operators, motivating skill-level shared autonomy via \textbf{IMCopilot} for improved performance and adaptability.

\subsection{Imitation learning with Force/Tactile Observation}
Prior imitation learning for robotic manipulation relies primarily on vision and proprioception, learning policies that map observed states to actions without direct force or tactile feedback for contact perception. To address this gap, recent work integrates force or tactile observation into imitation learning frameworks for contact-rich manipulation tasks~\cite{zhu2025touch, zhang2025ta, yu2025forcevla, huang2025tactile, he2025foar, huang20243d, xue2025reactive}.Among these, RDP~\cite{xue2025reactive} incorporates force or tactile signals as conditional inputs within a fast–slow architecture to refine actions, demonstrating strong performance in contact-rich manipulation. TA-VLA~\cite{zhang2025ta} investigates torque-aware Vision-Language-Action models, exploring architectural mechanisms for integrating torque signals to improve contact-rich manipulation. ForceVLA~\cite{yu2025forcevla} extends VLAs to contact-rich manipulation by treating force sensing as a first-class modality and introducing a force-aware Mixture-of-Experts fusion module for integrating visual-language representations with real-time force feedback. Tactile-VLA~\cite{huang2025tactile} introduces a multimodal framework that integrates vision, language, action, and tactile sensing with hybrid position–force control and tactile-aware reasoning, enabling adaptive contact-rich manipulation and zero-shot generalization with minimal demonstrations. However, most prior studies are limited to two-finger grippers~\cite{wu2026pragmatic, cui2025end, sapkota2025vision, fang2025dexop}, where tactile and force signals demonstrably improve contact-rich manipulation. 
In contrast, large-scale integration on high-DoF dexterous hands remains largely unexplored due to hardware complexity and dataset scarcity, while the high-dimensional action space poses additional challenges for VLA-based policies.

\section{Method}

\begin{figure}
    \centering
    \includegraphics[width=0.95\linewidth]{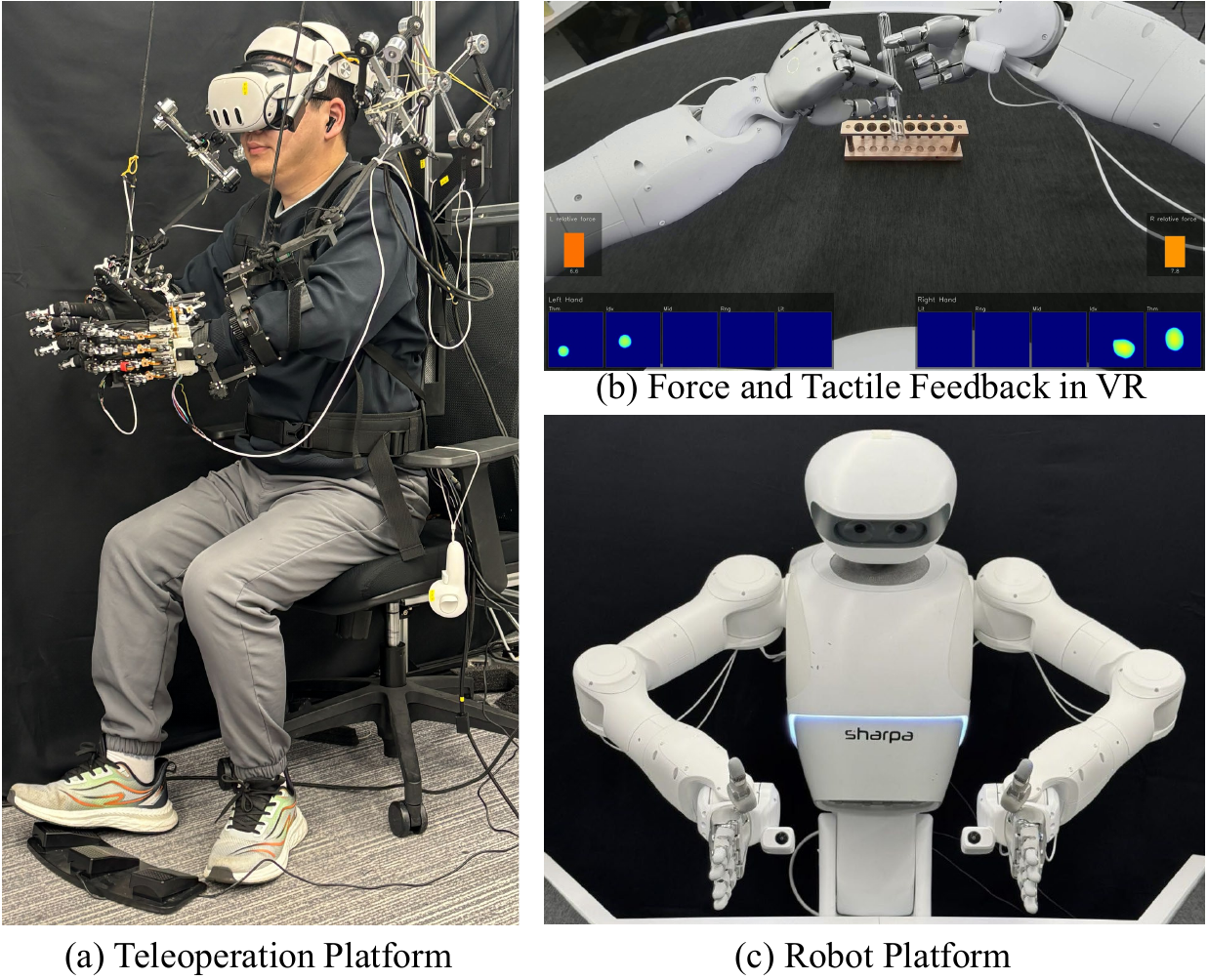}
    \caption{System Overview. (a) The teleoperation system, with exoskeletons, a VR headset, and foot pedals. (b) The VR view, integrating the robot's camera stream with real-time force and tactile feedback overlays. (c) The robot platform used for executing contact-rich tasks.}
    \label{fig:robot_and_data_acq}
    \vspace{-15pt}
\end{figure}

\begin{figure*}
    \centering
    \includegraphics[width=0.95\linewidth]{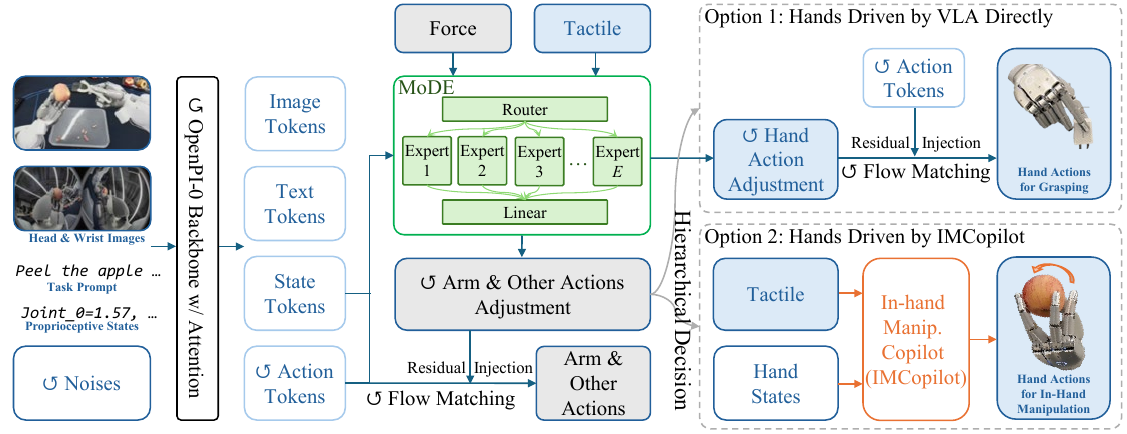}
    \caption{\textbf{Overview of MoDE-VLA.}
    Left: the OpenPI-0 backbone encodes visual, linguistic, proprioceptive, and noisy action inputs into token sequences.
    Center: the Mixture-of-Dexterous-Experts (MoDE) VLA ingests force and tactile observations, routes them through sparse experts, and produces modality-specific residual corrections---force-guided adjustments for arm actions and tactile-guided adjustments for hand actions.
    Right: a hierarchical decision mechanism selects between two options at each timestep: \emph{Option~1}, where hand actions are generated by the VLA with MoDE tactile refinement via flow matching, and \emph{Option~2}, where the RL-trained IMCopilot directly governs hand actions based on hand proprioceptive states. In both cases, arm and other actions are produced by the VLA with MoDE force refinement.}
    \label{fig:pipeline}
\end{figure*}

\subsection{Overview}

As discussed in \prettyref{sec:intro}, extending VLAs to contact-rich dexterous manipulation faces three key bottlenecks: (1) acquiring high-quality demonstrations for high-DoF systems, (2) covering qualitatively distinct manipulation phases with a single policy, and (3) integrating force and tactile modalities into a pretrained VLA backbone.
Our framework addresses these bottlenecks through two synergistic components.
\textbf{IMCopilot} (\prettyref{sec:method_rl}) provides a set of RL-trained atomic in-hand manipulation skills that serve a dual purpose: during data collection, they act as a shared-autonomy copilot that assists human operators through challenging in-hand phases (\emph{bottleneck 1}); during autonomous execution, the VLA invokes them as callable low-level primitives, forming a hierarchical manipulation policy (\emph{bottleneck 2}).
\textbf{MoDE-VLA} (\prettyref{sec:method_moe}) establishes a dedicated fusion pathway that integrates force and tactile information into the VLA's action expert via sparse expert routing and residual injection, achieving contact-aware control without degrading pretrained capabilities (\emph{bottleneck 3}).

\subsection{Robot Platform and Data Acquisition System}
\label{sec:method_system}

\paragraph{Robot platform}
\prettyref{fig:robot_and_data_acq} (c) illustrates our robotic platform, \texttt{SharpaNorth}\footnote{\url{https://www.sharpa.com/pages/north}}, which features dual 7-DoF robotic arms equipped with 22-DoF \texttt{SharpaWave}\footnote{\url{https://www.sharpa.com/pages/wave}} dexterous hands.
Each fingertip embeds a vision-based tactile sensor that provides 6-DoF force and wrench readings via the hand's integrated API, following the method proposed in~\cite{10342722}.
The robot additionally integrates a 2-DoF neck, a 3-DoF upper body, and a 2-DoF lower body (waist).
With the lower body fixed during our experiments, the system actively operates with a total of $(7+22)\times2+2+3=63$ DoFs.
For observation, two head-mounted RGB cameras provide stereo vision and two wrist-mounted fish-eye cameras capture close-range views.
Crucially, the system also records joint torque readings from both arms and 6-DoF force/wrench signals from all ten fingertip tactile sensors---these two streams constitute the \emph{force} and \emph{tactile} modalities that MoDE will later fuse into the VLA (\prettyref{sec:method_moe}).

\paragraph{Teleoperation system}
\prettyref{fig:robot_and_data_acq} (a) and (b) depict our data acquisition system, comprising an upper-body exoskeleton for arm kinematics, a pair of exoskeleton gloves for hand articulation, and a VR headset for immersive visual feedback.
The exoskeleton continuously tracks human arm and finger joint positions, which directly maps human configurations to the robot's action space.
The VR headset provides stereoscopic vision streamed from the robot's head cameras.
To give the operator a sense of contact state, data from the robot's joint torque sensors and fingertip tactile arrays are rendered as intuitive visual overlays within the VR.
The exoskeleton gloves also provide vibrotactile feedback to each fingertip when the robotic fingertips make contact with an object.
The wrist camera streams are not displayed to the operator to prevent visual clutter, but are recorded as standard observations for downstream policy training.
For more details, please refer to the supplementary video.

\paragraph{From teleoperation to shared autonomy}
This system achieves efficient data acquisition for standard pick-and-place tasks, as well as more complex maneuvers requiring bimanual coordination (e.g., transferring an object between hands).
Yet for the most demanding apple peeling task, even experienced operators found it infeasible to perform its prerequisite---stable, continuous in-hand rotation through direct teleoperation.
We thus introduce an RL-augmented shared-autonomy mechanism: the operator seamlessly triggers pre-trained hand skills (e.g., in-hand rotation) via foot pedals, while retaining full control of gross arm motions through the exoskeleton.
This hybrid setup significantly reduces operator cognitive load and yields high-quality demonstration trajectories that would otherwise be unobtainable.
The formulation and training of these skills are detailed next.

\subsection{Learning In-hand Manipulation Copilots}
\label{sec:method_rl}

\paragraph{Skill repertoire}
IMCopilot comprises a small set of atomic in-hand manipulation primitives
(i) \emph{stable grasp maintenance}, which keeps the object firmly held under external perturbations, and (ii) \emph{in-hand object rotation} around a specified axis (e.g., clockwise rotation of the apple about the vertical axis).
These primitives are designed to be composable: the higher-level decision-maker---whether a human operator or the VLA---selects \emph{which} skill to invoke and \emph{when}, while IMCopilot handles the low-level finger coordination.

\paragraph{RL training}
Each skill is trained using Proximal Policy Optimization (PPO)~\cite{schulman2017ppo} within the IsaacLab~\cite{mittal2025isaaclab} simulation environment, employing an asymmetric actor-critic architecture with teacher-student distillation~\cite{qi2022hora}.
To ensure the policy begins from physically feasible contact states, initial hand joint positions are sampled around a default pose and object positions are drawn uniformly within a predefined workspace; only configurations yielding stable grasps are accepted.
During the teacher phase, both actor and critic receive privileged information $\mathbf{e}_t$ (object pose, velocities, mass, center of mass, and friction coefficients) at time $t$ , which is encoded into a compact latent embedding and concatenated with the standard observation $\mathbf{o}_t$.
The student policy learns to regress this embedding directly from $\mathbf{o}_t$, enabling deployment without privileged state access.
The observation $\mathbf{o}_t$ comprises a 3-step history of proprioception, fingertip contact forces, and the target rotation axis.
The policy outputs relative joint position offsets $\mathbf{a}_t = \Delta\theta_t$, integrated as $\mathbf{q}_t = \mathbf{q}_{t-1} + \lambda_{\text{scale}}\Delta\theta_t$ and tracked by low-level PD controllers.
To facilitate zero-shot sim-to-real transfer, we apply domain randomization over object scale, mass, friction, center-of-mass offset, gravity, and PD gains.

The reward function $r = \lambda_{\text{rot}}r_{\text{rot}} + \lambda_{\text{vel}}r_{\text{vel}} + \lambda_{\text{work}}r_{\text{work}} + \lambda_{\text{torq}}r_{\text{torq}} + \lambda_{\text{diff}}r_{\text{diff}}$ encourages angular velocity around the target axis ($r_{\text{rot}}$) while penalizing undesired linear velocity ($r_{\text{vel}}$), excessive joint work ($r_{\text{work}}$), torque ($r_{\text{torq}}$), and joint deviation ($r_{\text{diff}}$) to ensure task progress stability.

\paragraph{Dual role: from teleoperation assistant to autonomous primitive}
IMCopilot plays a unified role across both the data collection and autonomous execution phases, as introduced in \prettyref{sec:intro}.
During data collection, the human operator triggers IMCopilot via foot pedals whenever in-hand manipulation is required, addressing the \emph{data acquisition bottleneck}.
During inference, the same skills are invoked by the VLA itself: the VLA's output action vector includes a scalar trigger signal $c \in [0, 1]$ alongside the standard arm and hand actions. $c > 0.5$ activates IMCopilot to execute in-hand rotation,
while arm control remains with the VLA, addressing the \emph{multi-skill learning challenge}.

\subsection{Fusing Force and Tactile into VLA with MoDE}
\label{sec:method_moe}

Contact-rich dexterous manipulation is fundamentally governed by contact dynamics, yet the VLA backbone was pretrained predominantly on parallel-gripper data where force feedback is absent.
Na\"ively appending force and tactile readings to the state vector treats them as just another input channel, ignoring two critical distinctions: these modalities carry different physical semantics (arm-level torques vs.\ fingertip contact patterns), and they evolve at different temporal scales than visual and language tokens.
As shown in prior work~\cite{yu2025forcevla, higuera2024sparsh}, such conflation degrades rather than enhances performance.
These observations motivate three design principles for MoDE-VLA: (i) a \emph{dedicated pathway} that processes force and tactile information separately from the pretrained backbone, (ii) \emph{modality-aware routing} that respects the distinct physical semantics of each signal, and (iii) \emph{residual injection} that contributes contact-aware corrections without overwriting the backbone's pretrained knowledge.
\prettyref{fig:pipeline} shows the overview of MoDE-VLA.

\subsubsection{VLA Backbone}

Our architecture extends a pretrained VLA backbone ($\pi_0$~\cite{black2024pi_0}) with the MoDE module.
The backbone consists of three components:
a SigLIP~\cite{zhai2023sigmoid} vision tokenizer (So400m/14),
a PaliGemma vision-language model (Gemma-3B~\cite{beyer2024pali}),
and
a lightweight action expert (Gemma-300M).
Given $K$ camera images $\{I_k\}_{k=1}^{K}$, a tokenized language instruction $\ell$, robot proprioceptive state $\mathbf{s} \in \mathbb{R}^{d_s}$, and a noisy action sequence $\mathbf{x}_t \in \mathbb{R}^{H \times d_a}$ (where $H$ is the action horizon and $d_a$ the action dimension), the model predicts a velocity field $\mathbf{v}_\theta(\mathbf{x}_t, t)$ that transports a noise distribution toward the target action distribution.
The training objective minimizes the flow-matching loss~\cite{lipman2022flow}:
\begin{equation}
\mathcal{L}_{\text{FM}} = \mathbb{E}_{t, \mathbf{x}_0, \boldsymbol{\epsilon}} \left[ \left\| \mathbf{v}_\theta(\mathbf{x}_t, t) - (\boldsymbol{\epsilon} - \mathbf{x}_0) \right\|^2 \right],
\label{eq:flow_matching}
\end{equation}
where $\mathbf{x}_t = t \cdot \boldsymbol{\epsilon} + (1 - t) \cdot \mathbf{x}_0$ linearly interpolates between clean actions $\mathbf{x}_0$ and noise $\boldsymbol{\epsilon} \sim \mathcal{N}(\mathbf{0}, \mathbf{I})$, with $t \sim \text{Beta}(1.5, 1)$.

Internally, the input is organized into a \emph{prefix} (vision, language, and proprioceptive state tokens) and a \emph{suffix} (noisy actions with timestep embeddings,).
The PaliGemma backbone processes the prefix, and the action expert processes the suffix; they share key-value pairs through cross-attention across all transformer layers.
At inference time, actions are generated by integrating the learned velocity field from $t{=}1$ (pure noise) to $t{=}0$ (clean action) via Euler's method with $N$ steps:
\begin{equation}
\mathbf{x}_{t+\Delta t} = \mathbf{x}_t + \Delta t \cdot \mathbf{v}_\theta(\mathbf{x}_t, t), \quad \Delta t = -\tfrac{1}{N}.
\end{equation}
The full action vector is $\mathbf{a} = [\mathbf{a}_{\text{arm}};\; \mathbf{a}_{\text{hand}};\; \mathbf{a}_{\text{other}}]$, where $\mathbf{a}_{\text{other}}$ includes waist actions and the IMCopilot trigger signal $c$.

\subsubsection{Force and Tactile Token Construction}

The \emph{force} signal $\mathbf{f} \in \mathbb{R}^{d_f}$ consists of joint torque readings from both robotic arms ($d_f = 14$, i.e., 7 joints $\times$ 2 arms).
These readings reflect arm-level contact forces---for example, the resistance encountered when the peeler cuts into an apple.
The \emph{tactile} signal $\mathbf{g} \in \mathbb{R}^{d_g}$ aggregates 6-DoF force and wrench measurements from all ten fingertip tactile sensors ($d_g = 60$, i.e., 5 fingers $\times$ 6 dimensions $\times$ 2 hands).
These readings capture fingertip-level contact patterns---for example, detecting slip onset or changes in grasp stability during object rotation.
The physical distinction between these two modalities---arm-level torques versus fingertip contact---is central to our fusion strategy and will directly inform the output routing in \prettyref{eq:residual}.

Each modality is projected into the PaliGemma embedding space of width $d_{\text{pali}}$ via a learned linear layer:
$\mathbf{z}_f = W_f \mathbf{f} + \mathbf{b}_f \in \mathbb{R}^{d_{\text{pali}}}$ and $\mathbf{z}_g = W_g \mathbf{g} + \mathbf{b}_g \in \mathbb{R}^{d_{\text{pali}}}$.
Since each raw reading is a single-frame snapshot while the action horizon spans $H$ future timesteps, we replicate each embedding $H$ times and add sinusoidal step-positional encodings to produce temporally-indexed token sequences:
\begin{equation}
\tilde{\mathbf{z}}_f^{(h)} = \mathbf{z}_f + \text{PE}_{\text{sin}}(h), \quad
\tilde{\mathbf{z}}_g^{(h)} = \mathbf{z}_g + \text{PE}_{\text{sin}}(h), \quad h = 1, \ldots, H.
\end{equation}
This yields $\tilde{\mathbf{Z}}_f, \tilde{\mathbf{Z}}_g \in \mathbb{R}^{H \times d_{\text{pali}}}$.
The replication serves an important design purpose: it creates $H$ independent slots so that the downstream MoE router can assign different experts to different timesteps along the action horizon---for instance, activating a contact-onset specialist for the first few steps and a steady-state force-tracking specialist for later steps.

\subsubsection{Mixture-of-Dexterous-Experts Module}
\label{sec:mode}

The MoDE block takes three streams of information---the backbone's contextual output, the current denoising state, and the force/tactile tokens---and produces contact-aware corrections to the predicted actions.
First, the module concatenates four token sequences along the sequence dimension:
\begin{equation}
\mathbf{Z}_{\text{in}} = \left[ \mathbf{Z}_{\text{prefix}} \;\|\; \mathbf{Z}_{\text{suffix}} \;\|\; \tilde{\mathbf{Z}}_f \;\|\; \tilde{\mathbf{Z}}_g \right] \in \mathbb{R}^{(S_p + 3H) \times d_{\text{pali}}},
\end{equation}
where $\mathbf{Z}_{\text{prefix}} \in \mathbb{R}^{S_p \times d_{\text{pali}}}$ represents a total number of $S_p$ tokens in the PaliGemma backbone's output, and $\mathbf{Z}_{\text{suffix}} \in \mathbb{R}^{H \times d_{\text{pali}}}$ is the action expert's suffix output corresponding to the $H$-step action horizon.
A self-attention layer processes this concatenated sequence, allowing the force and tactile tokens to attend simultaneously to the visual-linguistic context
and the denoising state.

The attended force and tactile tokens are then passed through a token-level Mixture-of-Experts layer comprising $E$ expert MLPs with top-$k$ scatter routing~\cite{shazeer2017outrageously, riquelme2021scaling}.
The key motivation for using sparse MoE, rather than a single shared MLP, is that contact-rich manipulation involves qualitatively different regimes---free-space reaching, initial contact, stable grasping, and dynamic in-hand rotation---each demanding distinct force-to-action mappings.
Sparse routing allows different experts to specialize in different regimes and different joint groups (e.g., arms vs.\ fingers), dynamically allocating network capacity based on the current manipulation phase without increasing per-token computation.

\subsubsection{Residual Injection and Action Generation}

The MoE layer outputs refined force tokens $\mathbf{Z}_f \in \mathbb{R}^{H \times d_{\text{pali}}}$ and tactile tokens $\mathbf{Z}_g \in \mathbb{R}^{H \times d_{\text{pali}}}$.
These are injected as residual corrections to the backbone's action prediction through modality-specific projection heads:
\begin{equation}
\mathbf{v}_\theta(\mathbf{x}_t, t) = \left[ W_1 \left( \mathbf{Z}_f + \mathbf{Z}_{\text{suffix}} \right) \;\|\; W_2 \left( \mathbf{Z}_g + \mathbf{Z}_{\text{suffix}} \right) \right],
\label{eq:residual}
\end{equation}
where $W_1$, $W_2$ are two separate linear projections layers for arm and hand actions.

This design offers two key advantages.
First, the residual structure ensures that MoDE functions strictly as a \emph{refinement} over the base VLA prediction. When force and tactile signals carry little information (e.g., during free-space motion), the correction naturally diminishes toward zero, preserving the backbone's robust pretrained behavior.
Second, the modality-specific routing avoids cross-contamination: arm-level torque information does not interfere with finger control, and vice versa, respecting the distinct physical semantics of each signal.
Note that \prettyref{eq:residual} describes action generation when the VLA retains full control (Option~1 in \prettyref{fig:pipeline}).
When IMCopilot is triggered ($c > 0.5$, Option~2), hand actions $\mathbf{a}_{\text{hand}}$ are overridden by it.

\section{Experiments}

\begin{figure*}
    \centering
    \includegraphics[width=0.95\linewidth]{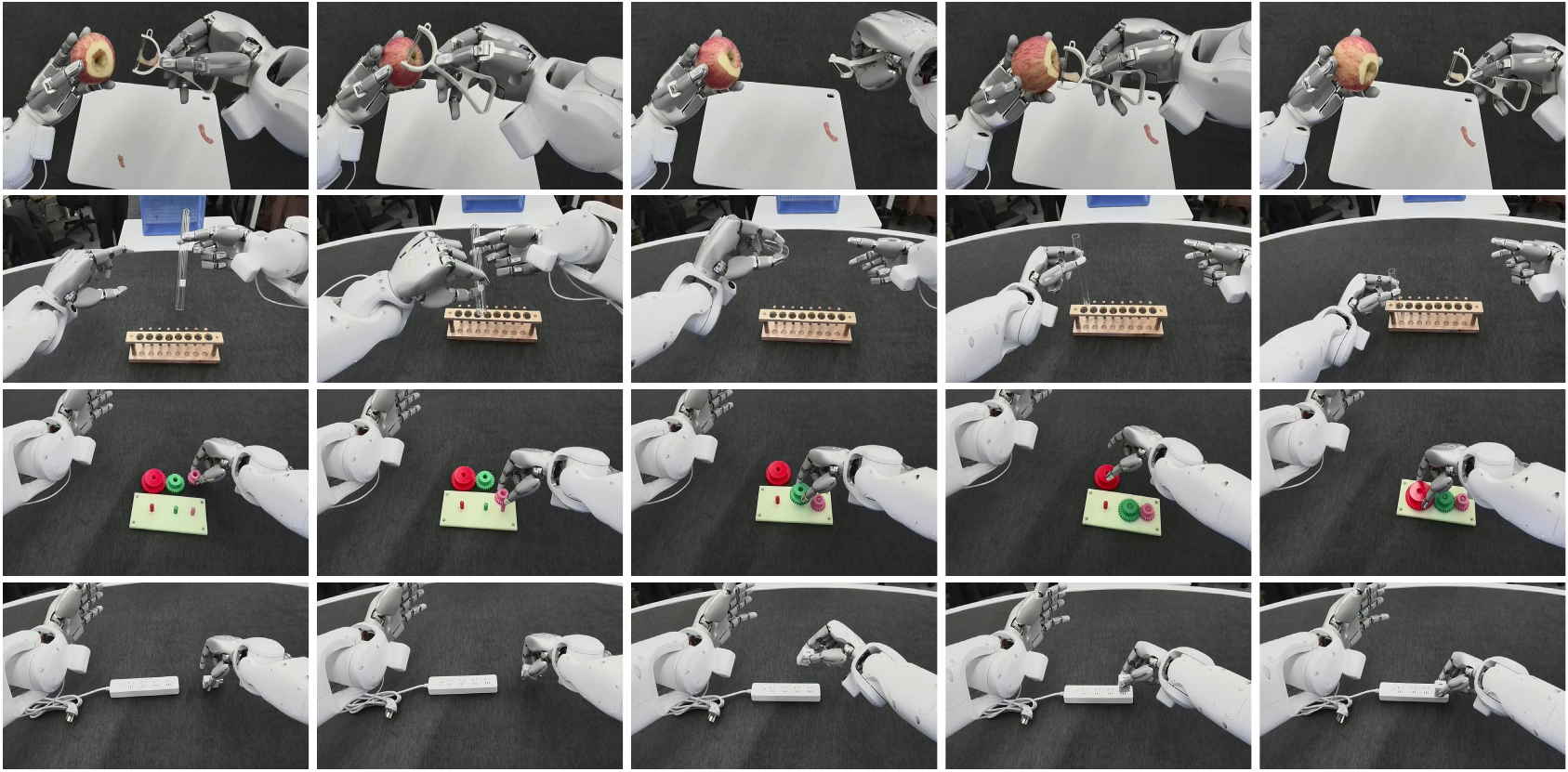}
    \caption{Illustration of the four evaluation tasks (rows): Apple Peeling, Tube Rearranging, Gear Assembling, and Charger Plugging. Each row shows five key frames of task execution from left to right.}
    \label{fig:tasks}
\end{figure*}

In this section, we conduct several experiments on different contact-rich manipulation tasks to evaluate the effectiveness of our method. We structure our analysis around the following research questions:

\noindent \textbf{Q1:} How well is the effectiveness of IMCopilot and force tactile feedback in data acquisition system?

\noindent \textbf{Q2:} How does MoDE-VLA perform compared to baselines?

\noindent \textbf{Q3:} How much does each component of MoDE-VLA contribute to its overall performance?

\subsection{Experimental Setup}

\textbf{Implementation Details}.
The robot platform and the corresponding data acquisition system have been discussed in \prettyref{sec:method_system}.
We initialize the VLA backbone from the pretrained $\pi_0$ checkpoint~\cite{black2024pi_0}.
The MoDE module and the related projection layers are initialized from scratch.
We use $E=8$ expert MLPs in the MoDE module with top-$k=1$ routing.
The action horizon is $H=50$ steps, and we denoise with $N=10$ Euler steps at inference.
Images from three camera views (base, left wrist, right wrist) are encoded at $224 \times 224$ resolution, yielding $256$ patch tokens per view ($768$ total image tokens).
We train the network for 60,000 steps with the AdamW optimizer with cosine learning rate decay and color jitter augmentation on input images.

\textbf{Task Definition}.
We evaluate on 4 contact-rich and difficult tasks (\prettyref{fig:tasks}).
Gear Assembling and Charger Plugging are single-armed tasks that require force feedback to regulate insertion forces, while Apple Peeling and Tube Rearranging are dual-armed tasks that require tight bimanual coordination. Details of each task:
    \textbf{Apple Peeling:} The robot holds a peeler in the right hand and an apple in the left hand, executing repeated peel-and-rotate cycles until a complete ring of peel is removed from the top of the apple.
    \textbf{Tube Rearranging:} The robot picks up an inverted tube from a rack by the right hand, swaps to the left hand, and inserts it into the designated slot.
    \textbf{Gear Assembling:} The robot sequentially picks up three gears from the table and inserts each onto its corresponding gear shaft.
    \textbf{Charger Plugging:} The robot grasps a charger from the table and plugs it into a power strip.
Note that we use IMCopilot to control the left hand in Peeling Apple and keep using VLA to directly drive the hands in the other tasks.

\textbf{Evaluation Protocol}.
We adopt two metrics to evaluate task performance.
The primary metric is the Success Rate (\textbf{SR}), defined as the fraction of trials of full success. %
For Apple Peeling, we additionally report the Peel Completion Ratio (\textbf{PCR}), defined as the fraction of the target apple surface successfully peeled, discretized into 25\% increments, which better captures partial progress in this cyclic task.
All methods are evaluated over 20 trials per task.
We compare against the base backbone model, $\pi_0$~\cite{black2024pi_0}.

\begin{table}
    \centering
    \setlength{\tabcolsep}{5pt}
    \begin{tabular}{lcc}
        \toprule
        \textbf{Object} & \textbf{Teleoperation} & \textbf{IMCopilot} \\
        \midrule
        Ping-pong ball  & 3/30 (10\%)  & 25/30 (83\%) \\
        Tennis ball     & 20/30 (67\%)  & 28/30 (93\%) \\
        Apple           & 8/30 (27\%)  & 27/30 (90\%) \\
        \midrule
        \textbf{Overall}& 31/90 (34\%)  & \textbf{80/90 (89\%)} \\
        \bottomrule
    \end{tabular}
    \caption{Successful rate of in-hand manipulation, in the case of teleoperation or using IMCopilot.}
    \label{tab:imcopilot_rotation}
    \vspace{-20pt}
\end{table}

\begin{table*}
    \centering
    \setlength{\tabcolsep}{5pt}
    \begin{tabular}{l cc c c c c}
        \toprule
        \multirow{2}{*}{\textbf{Method}}
          & \multicolumn{2}{c}{\textbf{Apple Peeling}}
          & \textbf{Tube Rearr.}
          & \textbf{Gear Assem.}
          & \textbf{Charger Plug.}
          & \textbf{Average}\\
        \cmidrule(lr){2-3}
          & SR $\uparrow$ & PCR $\uparrow$
          & SR $\uparrow$ & SR $\uparrow$ & SR $\uparrow$ & SR $\uparrow$\\
        \midrule
        $\pi_0$~\cite{black2024pi_0}    & / & 8\% & 15\%& 40\%& 5\%& 15\%\\
        \textbf{Ours}        & \textbf{30\%} & \textbf{73\%} & \textbf{30\%} & \textbf{60\%} & \textbf{15\%} & \textbf{34\%}\\
        \midrule
        \multicolumn{7}{l}{\textit{Ablations}} \\[2pt]
        ~~~w/o Force                    & 15\% & 48\% & 25\%& 45\%& 5\%& 23\%\\
        ~~~w/o Tactile                  & \textbf{30\%} & 70\% & 15\%& 45\%& 15\%& 26\%\\
        ~~~w/o IMCopilot$^\dagger$      & / & 25\% & -- & -- & -- & -- \\
        \bottomrule
        \multicolumn{7}{l}{\footnotesize /: Task failure (0\% SR). \quad --: Not applicable. \quad $^\dagger$IMCopilot is only used in Apple Peeling.}\\
    \end{tabular}
    
    \caption{
    Quantitative results across four tasks, against the baseline method $\pi_0$ and ablation studies.
    SR: Success Rate (\%). PCR: Peel Completion Ratio (\%, Apple Peeling only).
    Average SR is the mean success rate across all four tasks.
    }
    \label{tab:main_results}
    \vspace{-20pt}
\end{table*}

\subsection{Performance of Data Acquisition System (Q1)}
\textbf{Force and tactile feedback improve teleoperation efficiency and reliability}.
We evaluate how force and tactile visual feedback affect data collection efficiency and demonstration quality across 4 \emph{contact-rich} tasks. Without feedback, the operator relies solely on stereo VR images, making it difficult to judge grasp stability and contact forces, often leading to object slippage, overly strong gripping, or excessive insertion forces. With feedback enabled, these issues are significantly reduced. Quantitatively, take Gear Assembling as an example, without feedback the operator requires 75 minutes (including breaks) to complete 100 trials, achieving 85 successful demonstrations. With feedback, the operator completes 100 trials in 65 minutes with 93 successes, demonstrating improved data collection efficiency and success rate.

\textbf{IMCopilot enables reliable in-hand rotation that is difficult to achieve via plain teleoperation.}
Table~\ref{tab:imcopilot_rotation} summarizes the quantitative results. Teleoperation achieves an overall success rate of \textbf{34\% (31/90)}, whereas IMCopilot reaches \textbf{89\% (80/90)}. The gap is most pronounced for small objects: teleoperation succeeds in only \textbf{10\%} of trials for the ping-pong ball due to frequent drops during fingertip manipulation, while IMCopilot achieves \textbf{83\%} success through precise finger coordination. Teleoperation performs better for the tennis ball (\textbf{67\%}), where operators can stabilize the object against the palm, but struggles again for the apple (\textbf{27\%}), which is difficult to pivot within the hand. These results demonstrate the effectiveness and necessity of IMCopilot for data collection.

\subsection{MoDE-VLA Policy Performance (Q2)}
Table~\ref{tab:main_results} reports task success rates across all 4 tasks. MoDE-VLA achieves the highest average success rate (SR) of 34\%, outperforming the baseline by 19\%. The improvement is most pronounced on the two single-arm insertion tasks. On Gear Assembling and Charger Plugging, where success hinges on detecting contact onset and regulating insertion force in the final few millimeters of travel, MoDE-VLA outperforms $\pi_0$ by 20\% and 10\% respectively. We attribute this to the MoDE module: during contact, the internal routing mechanism activates force-specialized experts, enabling dynamic compliance modulation rather than relying on representations for free-space motion.

On the bimanual tasks, MoDE-VLA likewise outperforms the baseline. For Tube Rearranging, the higher success rate reflects improved inter-limb synchronization enabled by jointly processing force and tactile streams alongside shared visual context. For Apple Peeling, we report both SR 30\% and PCR 73\%. The PCR metric is particularly informative: baseline methods often begin peeling strokes correctly but fail to complete a full ring due to apple slippage or insufficient rotation, whereas MoDE-VLA's IMCopilot mechanism dispatches the RL rotation expert at the appropriate moment, enabling closed-loop ring completion. This is consistent with the finding in Q1 that conventional teleoperation achieves relatively low rotation success---the same structural difficulty that defeats direct teleoperation also defeats a VLA trained without an explicit rotation expert, and IMCopilot is the mechanism that bridges this gap at inference time.

\subsection{Ablation Studies (Q3)}
Table~\ref{tab:main_results} reports ablations of MoDE-VLA, where each variant removes a single design while keeping the others unchanged. Results show the sensing modalities and the MoDE module contribute independently to performance.

\textbf{Effect of force sensing}. Removing force input causes the largest single-component degradation, reducing the average SR by 11\%.
The drop in insertion tasks confirms that force signals provide the primary cue for contact onset detection and compliant insertion control. Without them, the model must infer contact state solely from visual observations, which is insufficient for millimeter-level insertion precision.
Additionally, during the apple-peeling task, the force-ablated model exhibits a higher tendency to fail to establish contact, resulting in peeling motions executed entirely in free space.

\textbf{Effect of tactile sensing}. Ablating the tactile visual stream results in an 8\% reduction in average SR, primarily manifesting as increased slip events during grasp-intensive phases. The tactile modality provides local fingertip deformation and contact state cues that are not captured by either wrist force/torque (F/T) sensing or egocentric RGB images alone. Without this signal, the model must rely on coarser contact proxies, which often fail when object compliance or surface friction varies across trials.
Interestingly, we find tactile ablation does not significantly degrade the apple-peeling success rate or PCR, as the knife is secured via a structural power grasp and the apple manipulation is delegated to IMCopilot, which already utilizes tactile information as input.

\textbf{Effect of IMCopilot}.
We attempted to train the VLA directly learn hand actions from IMCopilot demonstrations.
It leads to a substantial degradation in the Apple Peeling task, where the PCR drops from 73\% to 25\%.
This drop is primarily due to accidental dropping of the apple or failures to rotate it after the first successful peel.
These results highlight the importance of dedicated in-hand manipulation skills for executing the peel-and-rotate cycle. Without IMCopilot, the VLA has to directly coordinate high-DoF finger motions, which proves insufficient for reliable in-hand object rotation during peeling.

\section{Conclusions}

In this work, we extend VLA models to high-DoF, bimanual dexterous manipulation. 
We proposed a comprehensive hierarchical framework centered on two novel components. We introduced IMCopilot, which effectively bridges the teleoperation skill gap during data collection and provides robust, reactive low-level skills during autonomous execution. Furthermore, we developed MoDE module for VLA, which successfully fuses complex force and tactile feedback into pretrained VLA backbones through dedicated pathways and residual corrections. Extensive experimental evaluations across four challenging tasks—gear assembly, charger insertion, test tube handling, and apple peeling—confirm that our approach significantly outperforms the baseline method.

\bibliographystyle{IEEEtran}
\bibliography{references}

\end{document}